\documentclass[11pt]{article}

\usepackage[final]{acl}

\usepackage{times}
\usepackage{latexsym}

\usepackage[T1]{fontenc}

\usepackage[utf8]{inputenc}

\usepackage{microtype}

\usepackage{inconsolata}

\usepackage{graphicx}

\usepackage{amsmath}
\usepackage{amsfonts}
\usepackage{makecell}
\usepackage{xcolor}
\usepackage{tcolorbox}

\title{Training-Free Time Series Classification via In-Context Reasoning with LLM Agents}

\author{
{\bfseries Songyuan Sui, 
Zihang Xu, 
Xia Hu} \\
Rice University \\
\texttt{\{Songyuan.Sui, Zihang.Xu, Xia.Hu\}@rice.edu}
}

\begin{document}
\maketitle
\begin{abstract}
Time series classification (TSC) spans diverse application scenarios, yet labeled data are often scarce, making task-specific training costly and inflexible. Recent reasoning-oriented large language models (LLMs) show promise in understanding temporal patterns, but purely zero-shot usage remains suboptimal. We propose \textsc{FETA}, a multi-agent framework for training-free TSC via exemplar-based in-context reasoning. \textsc{FETA} decomposes a multivariate series into channel-wise subproblems, retrieves a few structurally similar labeled examples for each channel, and leverages a reasoning LLM to compare the query against these exemplars, producing channel-level labels with self-assessed confidences; a confidence-weighted aggregator then fuses all channel decisions. This design eliminates the need for pretraining or fine-tuning, improves efficiency by pruning irrelevant channels and controlling input length, and enhances interpretability through exemplar grounding and confidence estimation. On nine challenging UEA datasets, \textsc{FETA} achieves strong accuracy under a fully training-free setting, surpassing multiple trained baselines. These results demonstrate that a multi-agent in-context reasoning framework can transform LLMs into competitive, plug-and-play TSC solvers without any parameter training. The code is available at \url{https://github.com/SongyuanSui/FETATSC}.
\end{abstract}

\section{Introduction}

Time series classification (TSC) is a long-standing and fundamental task with wide applications ranging from healthcare \citep{wang2024medformermultigranularitypatchingtransformer}, finance \citep{liang2023improving} and human activity recognition \citep{ghorrati2024multi} to industrial monitoring \citep{farahani2025time}. Dominant solutions learn a trainable mapping from raw sequences to labels, either end-to-end with deep neural networks \citep{Karim_2019, zerveas2020transformerbasedframeworkmultivariatetime} or in two stages via representation learning that pretrains an encoder and then learns a downstream task-specific model \citep{eldele2021timeseriesrepresentationlearningtemporal, liu2023timesurlselfsupervisedcontrastivelearning}. While effective, these pipelines require dataset-specific training or fine-tuning, incurring nontrivial computational cost, hyperparameter tuning, and domain adaptation effort. Recently, Large Language Models (LLMs) have shown remarkable performance across a large variety of tasks beyond natural language processing (NLP) \citep{yang2023harnessingpowerllmspractice}. Studies that leverage LLMs for time series often still relies on additional training like fine-tuning \citep{xue2023promptcastnewpromptbasedlearning} and tokenization into patches \citep{nie2023timeseriesworth64, chang2025llm4ts}, limiting their generalization ability and plug-and-play usability. Moreover, the scarcity of labeled time series often makes such training insufficiently robust. For instance, \citet{zhou-etal-2025-merit} introduces a method to augment training data for time series representation learning, highlighting the difficulty of obtaining broadly generalizable models when data is limited.

More recently, reasoning-oriented LLMs such as GPT-o1 \citep{openai2024gpt4technicalreport}, DeepSeek-R1 \citep{deepseekai2025deepseekr1incentivizingreasoningcapability}, and Qwen3 \citep{yang2025qwen3technicalreport} have demonstrated the ability to capture temporal patterns such as trends, seasonality, and fluctuations, suggesting that they can process time series data through analogical reasoning \citep{chen2024conformalized, zhou2025enhancingllmreasoningtime}. However, existing studies also indicate that applying them in a purely zero-shot or fixed-prompt manner often yields suboptimal performance \citep{merrill2024languagemodelsstrugglezeroshot, liu2025evaluating1vs2}. This gap highlights an equally important but overlooked capability of LLMs in the time series domain: \textit{in-context learning}. Unlike conventional time series models, LLMs offer novel paradigms of learning and knowledge representation, particularly through in-context learning, which enables them to dynamically adapt their output based on exemplars without any parameter updates \citep{11018434}.

We thereby ask a different question: \emph{Instead of further expanding or retraining models, can we perform time series classification without any training?} We answer in the affirmative with \textbf{FETA} (training-\underline{F}ree tim\underline{E} series classifica\underline{T}ion with LLM \underline{A}gents), a multi-agent framework that replaces model fitting with exemplar-based, in-context reasoning. 

\textsc{FETA} decomposes a multivariate time series into channel-wise subproblems, retrieves a few similar labeled sequences per channel using Dynamic Time Warping (DTW) \citep{berndt1994using}, and leverages a reasoning LLM to compare the query against these exemplars, producing channel-level predictions with associated confidences. A confidence-weighted fusion then aggregates these outputs into the final decision. This design avoids pretraining or fine-tuning, improves efficiency by pruning irrelevant channels and controlling input length, and remains interpretable by grounding predictions in exemplars and confidence scores. The entire pipeline is built within a multi-agent framework, where each component operates independently yet contributes to a coherent, modular workflow.

To comprehensively evaluate the effectiveness of \textsc{FETA}, we conduct experiments across nine representative and challenging UEA benchmarks. The results demonstrate that \textsc{FETA} consistently achieves strong accuracy under a fully training-free setting, surpassing a variety of classical, deep learning, and representation learning baselines. These findings validate the effectiveness of our key architectural design: (i) decomposing multivariate sequences into channel-wise subproblems to focus reasoning on the most informative signals; (ii) retrieving structurally aligned exemplars to ground in-context reasoning; and (iii) aggregating per-channel decisions through confidence-weighted fusion. Together, these components enable robust and interpretable time series classification without any model training or parameter updates.

In summary, we propose \textsc{FETA}, a novel multi-agent collaboration framework for training-free time series classification via in-context reasoning with the following contributions:

\begin{itemize}
    \item We introduce the first training-free multi-agent framework for multivariate time series classification that leverages LLMs’ in-context reasoning ability, eliminating the need for pretraining, fine-tuning, or task-specific models.
    \item We design a multi-agent pipeline that integrates channel decomposition, exemplar retrieval, LLM in-context reasoning, and confidence-weighted aggregation, enabling efficient, interpretable, and modular classification.  
    \item We conduct extensive experiments on nine UEA benchmarks with both reasoning and non-reasoning LLMs, showing that \textsc{FETA} not only achieves state-of-the-art accuracy with reasoning LLMs but also delivers competitive results with smaller non-reasoning models, highlighting the effectiveness of our exemplar-based in-context learning approach. 
\end{itemize}

\section{Preliminaries}

In this section, we present the notations used throughout the paper, followed by the definitions of multivariate time series and the time series classification task. We then introduce the basic formulation of the multi-agent framework, which will be leveraged in the proposed method to enable training-free classification with large language models.

\subsection{Multivariate Time Series Definition}

A \emph{multivariate time series} is defined as
\begin{equation}
    \mathbf{X} = [\mathbf{x}_1, \mathbf{x}_2, \dots, \mathbf{x}_T] \in \mathbb{R}^{T \times C}, \quad \mathbf{x}_t \in \mathbb{R}^C,
\end{equation}
where $T$ is the number of time steps and $C$ is the number of variables (channels). 
Each time step vector is written as
\begin{equation}
    \mathbf{x}_t = [x_t^{(1)}, x_t^{(2)}, \dots, x_t^{(C)}]^\top,
\end{equation}
which denotes the measurements of all $C$ variables at time step $t$. This work considers datasets where each time series $\mathbf{X}$ is associated with a ground-truth label $y \in \mathcal{Y}$.

\subsection{Time Series Classification Task}

Given a time series $\mathbf{X} \in \mathbb{R}^{T \times C}$, 
the time series classification task seeks a mapping
\begin{equation}
    f: \mathbb{R}^{T \times C} \rightarrow \mathcal{Y},
\end{equation}
where $\mathcal{Y} = \{1, 2, \dots, K\}$ is the finite set of $K$ target classes. Traditional approaches parameterize $f$ as a trainable model $f_{\theta}$ and learn parameters $\theta$ by minimizing a supervised loss on labeled training data. In our setting, $f$ is instead implemented by a multi-agent framework that relies on LLMs' reasoning ability to directly predict labels without parameter training.

\subsection{Multi-Agent Framework Basics}
\label{sec:multiagent}

We define a multi-agent framework as a set of LLM-powered agents
\begin{equation}
    \mathcal{A} = \{A_1, A_2, \dots, A_M\},
\end{equation}
where each agent $A_m$ takes an input $I_m$ and produces an output
\begin{equation}
    O_m = A_m(I_m).
\end{equation}

Agents interact via a communication graph $\mathcal{G}_{\text{comm}} = (\mathcal{A}, \mathcal{E})$, where an edge $(A_i, A_j) \in \mathcal{E}$ indicates that the output of $A_i$ is passed to $A_j$. The final prediction $\hat{y} \in \mathcal{Y}$ is obtained collaboratively as
\begin{equation}
    \hat{y} = \mathcal{F}(\mathcal{A}, \mathcal{G}_{\text{comm}}),
\end{equation}
with $\mathcal{F}$ denoting the overall decision function induced by agent interactions.

\section{Related Work}
We review previous work from three main aspects: Time Series Classification, Large Language Models for Time Series Analysis, and Multi-Agent Collaboration.

\noindent\textbf{Time Series Classification.}
Time series classification is a long-standing task in time series analysis. Classical methods often rely on distance-based algorithms such as k-Nearest Neighbors (k-NN) \citep{lee2012nearest} with metrics like Euclidean distance or Dynamic Time Warping. With deep learning, a wide range of models have been proposed, including RNNs \citep{lai2018modelinglongshorttermtemporal}, LSTMs \citep{karim2017lstm}, CNNs \citep{luo2024moderntcn}, Transformers \citep{liu2021gated, le2024shapeformer}, and hybrid architectures \citep{khan2021bidirectional, zhang2024dual}. While effective, these models typically require dataset-specific training, limiting scalability. More recently, representation learning approaches pretrain encoders to map raw signals into embedding spaces for downstream tasks. Examples include TS-TCC \citep{eldele2021timeseriesrepresentationlearningtemporal} with contrastive temporal context, T-Loss \citep{franceschi2020unsupervisedscalablerepresentationlearning} with autoregressive prediction, TS2Vec \citep{yue2022ts2vecuniversalrepresentationtime} with hierarchical contrastive objectives, and Times-URL \citep{liu2023timesurlselfsupervisedcontrastivelearning} with unified self-supervised objectives. These methods generalize well across benchmarks but still require training a dedicated encoder and then fine-tuning or attaching a task-specific model like classifier, making the two-stage pipeline costly and less flexible.

\noindent\textbf{Large Language Models for Time Series Analysis.}
Recent advancements in LLMs lead to growing interest in their application to time series analysis. Early methods adapt pretrained LLMs to the textualized time series data by fine-tuning them with task-specific datasets \citep{xue2023promptcastnewpromptbasedlearning}. A more popular strategy is to convert raw time series into patches that serve as input tokens \citep{nie2023timeseriesworth64}. GPT4TS \citep{zhou2023fitsallpowergeneraltime} reprograms the input time series into text style representations and augments the context with declarative prompts. LLM4TS \citep{chang2025llm4ts} uses a two-stage fine-tuning strategy to align patched time series data with LLMs and the downstream tasks. InstructTime \citep{cheng2024advancingtimeseriesclassification} reframes time series classification as a multimodal instruction-following task by discretizing signals into tokens and fine-tuning a language model to generate textual labels. However, despite promising results, these approaches still rely on additional training, which incurs considerable computational cost. Moreover, while LLMs are inherently valued for their cross-task generality, most existing studies tend to yield time series forecasting models. Extending these models to other time series applications such as classification and anomaly detection requires extra adaptation or retraining, limiting their universality.

\begin{figure*}[t]
  \centering
  \includegraphics[width=\linewidth]{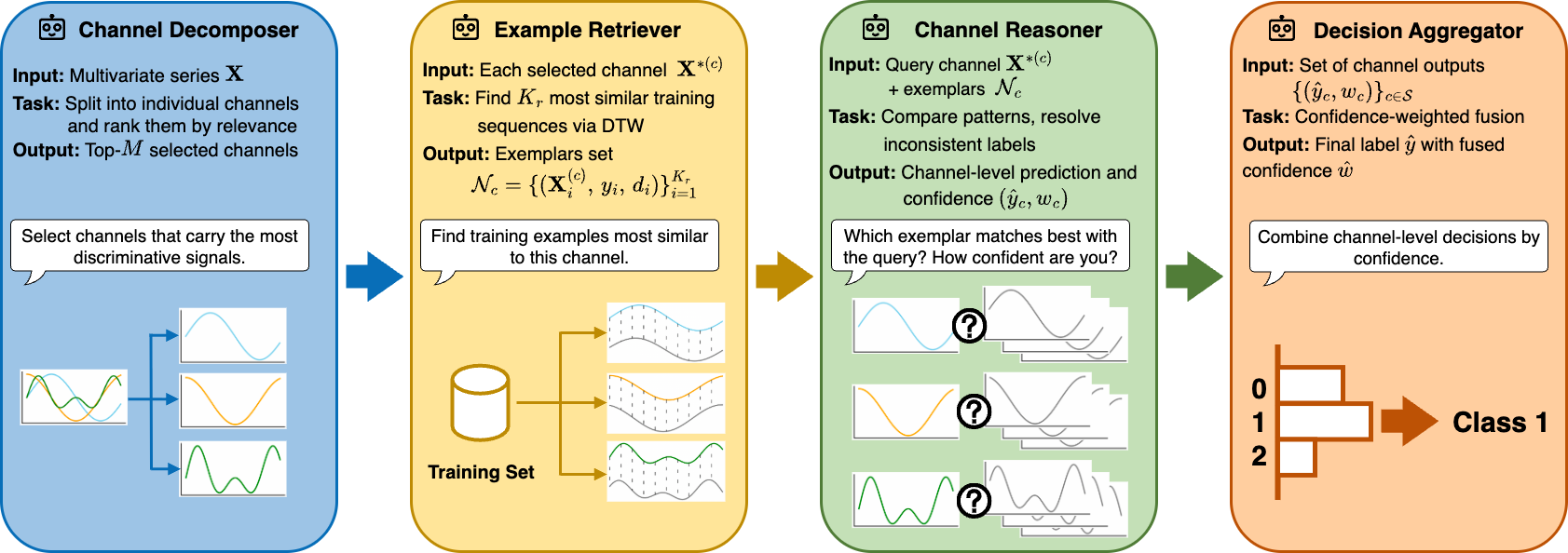}
  \caption{Overview of Our \textsc{FETA} Framework.}
  \label{fig:overview}
\end{figure*}

\noindent\textbf{Multi-Agent Collaboration.}
Recent studies enable multiple LLM-powered agents to address complex problems through extensive collaboration and collective intelligence \citep{wu2023autogen}. These approaches expand LLMs' reasoning capabilities beyond NLP domains. Research has applied multi-agent collaboration in programming \citep{xie2024magsqlmultiagentgenerativeapproach, wang2025macsqlmultiagentcollaborativeframework}, structured data processing \citep{wang2024chainoftableevolvingtablesreasoning, sui2025chainofqueryunleashingpowerllms}, and computer vision \citep{jiang2024multiagentvqaexploringmultiagent, li2025mccd}. Recently, researchers have also begun exploring multi-agent collaboration for time series analysis. TS-Reasoner \citep{ye2025domainorientedtimeseriesinference} introduces a multi-agent framework that leverages an LLM as a task decomposer and collaborates with specialized operators for multi-step time series inference, but it focuses on complex reasoning and relies heavily on predefined toolkits. MERIT \citep{zhou-etal-2025-merit} proposes a multi-agent framework to augment training data for time series representation learning, but still requires training a dedicated encoder, limiting plug-and-play applicability. ReasonTSC \citep{zhou2025enhancingllmreasoningtime} integrates typical time series analysis models with LLM-based reasoning to improve classification, yet still depends on well-trained base time series classifiers.

\section{Methodology}

\subsection{Overview}

We propose a multi-agent framework that decomposes multivariate time series classification into modular sub-tasks, with each agent responsible for a specific stage of the pipeline. As shown in Figure~\ref{fig:overview}, the framework consists of four specialized agents: \textbf{Channel Decomposer}, \textbf{Example Retriever}, \textbf{Channel Reasoner}, and \textbf{Decision Aggregator}. The Channel Decomposer isolates individual univariate channels from the multivariate input and filters out relatively irrelevant ones, ensuring that downstream reasoning focuses only on informative signals. This design enables the system to function in a fully training-free manner, leveraging nearest-neighbor evidence and lightweight in-context reasoning rather than model fine-tuning. Together, these agents form a coherent pipeline for training-free time series classification, which we describe in detail in the following sections.

\subsection{Channel Decomposer: Selection and Preprocessing}
\label{sec:decomposer}

The Channel Decomposer serves as the entry point of the entire agentic workflow. Time series classification spans a wide range of application scenarios, where sequence lengths can vary from a few dozen to several thousand time steps and the number of channels can range from a handful to thousands. Such heterogeneity poses significant challenges for direct LLM-based reasoning. The module therefore has two purposes: (1) enforcing channel independence, i.e., representing each input sequence using only one channel, which has been shown to preserve discriminative temporal patterns in Transformer-based architectures \citep{nie2023timeseriesworth64}; and (2) pruning irrelevant or redundant channels to reduce reasoning overhead, particularly when the original dimensionality is large. In this way, the module optimizes the input for downstream LLM-based reasoning by reducing noise and focusing on the most informative channels.

Formally, given $\mathbf{X}\in\mathbb{R}^{T\times C}$, the Channel Decomposer isolates channel-wise sequences
\[
    \mathbf{X}^{(c)} = [x^{(c)}_1, \dots, x^{(c)}_T], \quad c \in \{1, \dots, C\},
\]
and ranks channels by a fused, training-free relevance score computed on the training set. The goal is to identify the most informative channels, reduce redundancy, and ensure that downstream agents focus on dimensions most correlated with class distinctions. The following subsections detail this process.

\paragraph{Length normalization and per-sample $z$-normalization.}
Time series from different channels and samples often have varying lengths and scales. To make them comparable, we follow the preprocessing in \citep{dau2019ucrtimeseriesarchive} to normalize each channel independently:
\begin{equation}
    \tilde{\mathbf{X}}^{(c)} = \frac{\mathbf{X}^{(c)} - \mu}{\sigma},
\end{equation}
where $\mu$ and $\sigma$ denote the mean and standard deviation of $\mathbf{X}^{(c)}$.  
We then uniformly subsample each sequence to a fixed target length $L$ via uniform index mapping. This step prevents overly long sequences from inflating the LLM input size and degrading reasoning efficiency, while preserving global shape and trends through uniform subsampling.  
The resulting per-channel training set is denoted as
\[
    \mathcal{D} = \{(\tilde{\mathbf{X}}^{(c)}_i, y_i)\}_{i=1}^N.
\]
This preprocessing ensures that subsequent channel evaluation operates on standardized and comparable inputs.

\paragraph{Prototype-margin score (B).}
An informative channel should maximize inter-class separation while minimizing intra-class variance. To quantify this, we compute a prototype-margin score that compares inter-class separation against intra-class compactness, akin to Fisher discriminant ratio in classical feature selection \citep{gu2012generalizedfisherscorefeature}. Specifically, for each class $y\in\mathcal{Y}$ we compute the class centroid
\begin{equation}
    \boldsymbol{\mu}_y = \frac{1}{|\mathcal{D}_y|}\sum_{i\in\mathcal{D}_y}\tilde{\mathbf{X}}^{(c)}_i,
\end{equation}
the average within-class spread
\begin{equation}
    \mathrm{W} = \frac{1}{\sum_y|\mathcal{D}_y|}\sum_y\sum_{i\in\mathcal{D}_y}\|\tilde{\mathbf{X}}^{(c)}_i-\boldsymbol{\mu}_y\|_2,
\end{equation}
and the mean pairwise centroid distance
\begin{equation}
    \mathrm{B} = \frac{2}{K(K-1)}\sum_{y<y'}\|\boldsymbol{\mu}_y-\boldsymbol{\mu}_{y'}\|_2.
\end{equation}
The prototype-margin score is then
\begin{equation}
    B_c = \frac{\mathrm{B}}{\mathrm{W}+\varepsilon}, \quad \varepsilon>0.
\end{equation}
A higher $B_c$ indicates that channel $c$ better separates classes with compact clusters, making it more discriminative.

\paragraph{Approximate 1NN leave-one-out accuracy (C).}
Beyond cluster structure, we also evaluate how well each channel supports classification via nearest-neighbor retrieval. To this end, we adopt leave-one-out 1NN accuracy as a channel-level discriminability measure, which is a standard evaluation criterion in time series classification \citep{lines2015time}. Specifically, we use a probe subset of size $n_{\text{probe}}$ for efficiency. For each probed index $i$, we find
\begin{equation}
    j^\star = \arg\min_{j\neq i}\|\tilde{\mathbf{X}}^{(c)}_i-\tilde{\mathbf{X}}^{(c)}_j\|_2^2,
\end{equation}
and check if $y_{j^\star}=y_i$. The channel score is
\begin{equation}
    C_c = \frac{1}{n_{\text{probe}}}\sum_{i\in\text{probe}}\mathbf{1}[y_{j^\star}=y_i].
\end{equation}
Channels with higher $C_c$ demonstrate stronger standalone classification ability, making them valuable candidates.

\paragraph{Score fusion and ranking.}
Since $B_c$ emphasizes structural separation and $C_c$ emphasizes predictive accuracy, we combine them to obtain a balanced channel relevance score. After standardizing both via $z$-scores, we compute
\begin{equation}
    S_c = \alpha\,\mathrm{z}(B_c) + (1-\alpha)\,\mathrm{z}(C_c), \quad \alpha\in[0,1].
\end{equation}
Channels are then ranked by $S_c$, and the top-$M$ are selected for downstream reasoning. This fusion balances structural separability and predictive accuracy, ensuring that the selected channels provide meaningful evidence for downstream reasoning.

\medskip
In summary, the Channel Decomposer provides a lightweight, training-free mechanism to evaluate and select channels. By enforcing channel independence and pruning redundancy, the module yields a compact, informative subset of channels, reducing input dimensionality and streamlining downstream LLM reasoning.

\subsection{Example Retriever: In-Context Exemplar Selection}
\label{sec:retriever}

After selecting a compact set of informative channels, the next step is to provide each channel agent with concrete exemplars that can guide in-context reasoning. The motivation here is aligned with in-context learning \citep{liu2021makesgoodincontextexamples}: instead of training parameters, we supply the LLM with a small set of relevant examples, allowing it to adapt to the classification task by analogy. In the time series domain, this means retrieving similar labeled sequences from the training set so that the LLM can compare the query against concrete precedents. This design is supported by recent evidence that LLMs are capable of understanding fundamental temporal patterns such as trends, seasonality, and fluctuations \citep{zhou2025enhancingllmreasoningtime}, making it feasible to directly compare retrieved exemplars with the query sequence for classification. Meanwhile, labeled time series are often scarce; \citet{zhou-etal-2025-merit} resort to multi-agent data augmentation to compensate for limited training data in representation learning. In contrast, our retriever operates \emph{without} any model training or augmentation: only a handful of labeled exemplars are retrieved on-the-fly per query to ground reasoning.

Formally, for each selected channel $c \in \mathcal{S}$ and a test sequence $\mathbf{X}^{*(c)}$, the Example Retriever searches for its $K_r$ most similar training sequences using Dynamic Time Warping (DTW) \citep{berndt1994using}. The retrieved set is
\begin{equation}
    \mathcal{N}_c = \{(\mathbf{X}_i^{(c)}, y_i, d_i)\}_{i=1}^{K_r},
\end{equation}
where $d_i = \text{DTW}(\mathbf{X}^{*(c)}, \mathbf{X}_i^{(c)})$ denotes the alignment-based distance between the test and the $i$-th training sequence. DTW is particularly suitable for this role because it robustly accounts for local temporal distortions, variable speeds, and phase shifts, allowing two sequences with similar overall patterns but misaligned time indices to be matched effectively. This makes it a long-standing standard for time series similarity, and far more reliable than rigid measures such as Euclidean distance.

The retrieved exemplars $\mathcal{N}_c$ provide both labels $\{y_i\}$ and DTW distances $\{d_i\}$. Since the $K_r$ neighbors may disagree, we do not rely on majority voting; instead, we present them as diverse candidates and summarize their evidence (e.g., label histograms, neighbor-distance profiles, and representative exemplars) to form the in-context “demonstration set’’ for the Channel Reasoner.

In short, the Example Retriever enables training-free, interpretable classification by grounding each channel agent in nearest-neighbor evidence rather than learned representations. Leveraging DTW ensures that similarity reflects structural alignment rather than rigid pointwise matching, yielding robust, informative exemplars, even under label-scarce regimes.

\subsection{Channel Reasoner: Pattern-Based In-Context Reasoning}
\label{sec:reasoner}

The $K_r$ nearest neighbors for each channel provide concrete exemplars, but their labels may disagree; a naive majority vote can overlook temporal nuances that differentiate classes. The Channel Reasoner therefore leverages a reasoning LLM to perform in-context comparison of temporal patterns: given the query sequence and its retrieved exemplars, the agent contrasts trends, fluctuations, alignments, and other descriptive characteristics to adjudicate among conflicting candidates.

For each selected channel $c \in \mathcal{S}$, we instantiate an agent $A_c \in \mathcal{A}$. The agent is presented with the query $\mathbf{X}^{*(c)}$ and the retrieved exemplar set
\[
\mathcal{N}_c = \{(\mathbf{X}_i^{(c)}, y_i, d_i)\}_{i=1}^{K_r},
\]
where $d_i=\mathrm{DTW}(\mathbf{X}^{*(c)}, \mathbf{X}_i^{(c)})$ are alignment-based distances from the Example Retriever. Based on this input, the reasoning LLM outputs a channel-level decision
\begin{equation}
    O_c = (\hat{y}_c, w_c) = A_c\big(\mathbf{X}^{*(c)}, \mathcal{N}_c\big),
\end{equation}
where $\hat{y}_c \in \mathcal{Y}$ is the predicted label for channel $c$, and $w_c \in [0,1]$ is a confidence score explicitly generated by the model. Each channel agent operates independently, and all reasoners run in parallel without inter-agent communication, enabling scalable and efficient inference across high-dimensional time series.

Because we employ a reasoning LLM, the agent can internally incorporate descriptive statistics of the query $\mathbf{X}^{*(c)}$ (e.g., mean, variance, extrema, turning points) without explicit feature engineering. The prompt instructs $A_c$ to: (i) compare the query with each exemplar, using DTW distances $\{d_i\}$ as similarity evidence; (ii) identify the exemplars whose temporal patterns align best with the query; and (iii) justify the chosen label while providing a self-assessed confidence $w_c$. This converts raw nearest-neighbor evidence into a principled, interpretable decision rather than a heuristic vote.

In summary, the Channel Reasoner operationalizes exemplar-based in-context reasoning at the channel level. By combining retrieved neighbors with intrinsic descriptive cues of the query, the reasoning LLM produces both a label $\hat{y}_c$ and a confidence score $w_c$, which together form the structured outputs passed to the Decision Aggregator.

\subsection{Decision Aggregator: Confidence-Weighted Fusion}
\label{sec:aggregator}

The final stage aggregates channel-level predictions into a single decision. Because different channels may provide complementary or conflicting evidence, the aggregator combines labels \(\{\hat{y}_c\}_{c\in\mathcal{S}}\) and confidences \(\{w_c\}_{c\in\mathcal{S}}\) into a robust consensus via confidence-weighted fusion.

Given the candidate class set \(\mathcal{Y}\) and all agent outputs, we first discard invalid predictions (labels not in \(\mathcal{Y}\)). If all valid agents predict the same label, we enter \emph{consensus mode} and compute the fused confidence as
\[
\hat{w} \;=\; 1 - \prod_{c\in\mathcal{S}} (1 - w_c),
\]
which increases with both the number of agreeing agents and their confidences. The final decision is this common label \(\hat{y}\) with confidence \(\min(0.99,\hat{w})\).

If predictions are not unanimous, we switch to \emph{confidence-weighted mode}. Let the clipped confidence be
\(\tilde{w}_c=\min(\text{clip}_{\mathrm{hi}},\,\max(\text{clip}_{\mathrm{lo}},\,w_c))\)
with fixed bounds \(0<\text{clip}_{\mathrm{lo}}<\text{clip}_{\mathrm{hi}}<1\).
Each agent contributes \(\tilde{w}_c\) to its predicted class and a small smoothing weight \(\varepsilon>0\) to other classes:
\[
s_y \;=\; \sum_{c\in\mathcal{S}}
\begin{cases}
\tilde{w}_c, & \text{if } \hat{y}_c = y,\\
\varepsilon, & \text{otherwise}.
\end{cases}
\]
We then normalize scores and select
\[
\hat{y} \;=\; \arg\max_{y\in\mathcal{Y}} s_y, 
\qquad 
\hat{w} \;=\; \frac{s_{\hat{y}}}{\sum_{y\in\mathcal{Y}} s_y}.
\]
If no valid agent remains after filtering, the aggregator returns a null decision with zero confidence.

In short, the aggregator balances agreement and diversity: it boosts confidence under strong consensus and re-weights votes by calibrated confidence under disagreement. This yields a final decision that is robust and interpretable, reflecting the strength of cross-channel evidence.


\begin{table*}[t]
\centering
\renewcommand{\arraystretch}{1.2}
\resizebox{\textwidth}{!}{
\begin{tabular}{lccccccccccccc}
\hline
Dataset & DTW & \makecell{WEASEL\\+MUSE} & \makecell{MLSTM\\-FCNs} & TST & TS-TCC & T-Loss & TS2Vec & \makecell{Times\\-URL} & \makecell{DeepSeek\\R1} & MERIT & \makecell{FETA\\(LLaMA3.1)} & \makecell{FETA\\(DeepSeek R1)} & \makecell{FETA\\(Qwen3)} \\
\hline
AtrialFibrillation      & 22.0  & 33.3 & 26.7 & 26.7 & 26.7 & 26.7 & 26.7 & 26.7 & 20.0  & 33.3 & 33.3 & \underline{40.0}  & \textbf{46.7} \\
ERing                   & 13.3  & 13.3 & 13.3 & 13.3 & 13.3 & 13.3 & 13.3 & 13.3 & 13.3 & 13.3 & \textbf{24.4} & \underline{23.7} & 23.0  \\
EigenWorms              & 55.7  & \textbf{89.0} & 50.4 & 61.1 & 61.8 & 61.1 & 62.6 & 64.1 & 48.4 & \underline{65.6} & 57.3 & 60.3 & 61.1 \\
EthanolConcentration    & 29.3  & \textbf{43.0} & 37.3 & 32.3 & 33.1 & 32.7 & 33.8 & 34.2 & 21.7  & 35.0  & 28.5 & 36.2 & \underline{38.0}  \\
FingerMovements         & 55.0  & 49.0 & 58.0 & 56.7 & 58.3 & 58.3 & 60.0 & \underline{61.7} & 38.0  & \textbf{63.3} & 51.0  & 57.0  & 58.0  \\
HandMovementDirection   & 27.8  & 36.5 & 36.5 & 30.6 & 33.3 & 30.6 & 33.3 & 36.1 & 25.7 & \textbf{38.9} & 25.7 & 35.1 & \underline{37.8} \\
MotorImagery            & 39.0  & 50.0 & \underline{51.0} & 42.0 & 44.0 & 43.0 & 45.0 & 46.0 & 40.0  & 48.0 & 43.0 & 50.0  & \textbf{52.0} \\
SelfRegulationSCP2      & 48.3  & 46.0 & 47.2 & 50.6 & 52.2 & 51.7 & 53.3 & 54.4 & 45.0 & 55.6 & 46.7 & \textbf{56.7} & \underline{56.1} \\
StandWalkJump           & 33.3  & 33.3 & 6.7  & 36.7 & 40.0  & 36.7 & 40.0  & 43.3 & 40.0  & 46.7 & 40.0  & \textbf{60.0}  & \underline{53.3} \\
\hline
Average                 & 35.9 & 43.7 & 36.3 & 38.9 & 40.3 & 39.3 & 40.9 & 42.2 & 32.5 & 44.4 & 38.9 & \underline{46.6} & \textbf{47.3} \\
\hline
\end{tabular}
}
\caption{Comparison of classification accuracy (\%) across datasets. Baselines cover diverse categories. \textsc{FETA} variants are compared under the same evaluation. Best results per dataset are in \textbf{bold}, and second-best results are \underline{underlined}. \textsc{FETA} achieves
the highest accuracy, driven by multivariate series decomposition, DTW-aligned exemplars retrieval, channel-level in-context reasoning, and confidence-weighted aggregation.}
\label{tab:main_results}
\end{table*}

\begin{table}[t]
\centering
\renewcommand{\arraystretch}{1.2} 
\resizebox{\columnwidth}{!}{
\begin{tabular}{lccccc}
\hline
Dataset & FETA & \makecell{w/o\\Decomposer} & \makecell{w/o\\Retriever} & \makecell{w/o\\Reasoner} & \makecell{w/o\\Aggregator} \\
\hline
AtrialFibrillation & \textbf{40.0} & 20.0 & 33.3 & 33.3 & 20.0 \\
$\Delta$           & --             & \textcolor{red}{-20.0} & \textcolor{red}{-6.7} & \textcolor{red}{-6.7} & \textcolor{red}{-20.0} \\
ERing              & \textbf{23.7} & 19.7 & 20.7 & 21.1 & 21.5 \\
$\Delta$           & --             & \textcolor{red}{-4.0}  & \textcolor{red}{-3.0} & \textcolor{red}{-2.6} & \textcolor{red}{-2.2} \\
StandWalkJump      & \textbf{60.0} & 40.0 & 46.7 & 40.0 & 53.3 \\
$\Delta$           & --             & \textcolor{red}{-20.0} & \textcolor{red}{-13.3} & \textcolor{red}{-20.0} & \textcolor{red}{-6.7} \\
\hline
Average            & \textbf{41.2} & 26.6 & 33.6 & 31.5 & 31.6 \\
$\Delta$           & --             & \textcolor{red}{-14.6} & \textcolor{red}{-7.6}  & \textcolor{red}{-9.7}  & \textcolor{red}{-9.6} \\
\hline
\end{tabular}
}
\caption{Ablation study of \textsc{FETA} (DeepSeek R1). 
Each $\Delta$ row shows the accuracy drop relative to the full model. \textsc{FETA} achieves optimal performance through its modular design, with each component contributing significantly to accurate classification.}
\label{tab:ablation}
\end{table}

\section{Experiments}
In this section, we empirically evaluate the effectiveness of \textsc{FETA}.

\subsection{Experimental Setup}

\noindent\textbf{Datasets.}~~We evaluate on nine representative \emph{and challenging} datasets from the UEA classification archive \citep{bagnall2018ueamultivariatetimeseries}. This subset spans diverse application scenarios and exhibits substantial heterogeneity in sequence length, channel count, and number of classes. We adopt the official train/test splits and report classification accuracy. Notably, this selection is intentionally difficult: as shown in Table~\ref{tab:main_results}, most baselines obtain only modest average accuracy on these nine datasets, underscoring the challenge and leaving ample room for improvement.

\noindent\textbf{Baselines.}~~We compare our method against a various types of state-of-the-art baselines, including the distance-based method DTW \citep{berndt1994using}, pattern-based method WEASEL+MUSE \citep{schäfer2018multivariatetimeseriesclassification}, deep learning model MLSTM-FCNs \citep{Karim_2019}, transformer-based model TST \citep{zerveas2020transformerbasedframeworkmultivariatetime}, representation learning methods TS-TCC \citep{eldele2021timeseriesrepresentationlearningtemporal}, T-Loss \citep{franceschi2020unsupervisedscalablerepresentationlearning}, TS2Vec \citep{yue2022ts2vecuniversalrepresentationtime}, Times URL \citep{liu2023timesurlselfsupervisedcontrastivelearning}, and the multi-agent method MERIT \citep{zhou-etal-2025-merit}.

\noindent\textbf{Implementation Details.}~~We maintain the original training-test splits from the UEA archive. we adopt LLaMA~3.1, DeepSeek R1, and Qwen 3 as the backbone LLMs. Model configurations are detailed in Appendix~\ref{sec:appendix_parameters}. Prompts are dataset-agnostic, as shown in Appendix~\ref{sec:appendix_prompts}.

\subsection{Main Results}

Here, we compare \textsc{FETA} against diverse baselines on nine datasets. As shown in Table~\ref{tab:main_results}, all three \textsc{FETA} variants perform strongly despite requiring \textbf{no training}. FETA (Qwen3) achieves the best average accuracy at 47.3\%, surpassing strong classical (WEASEL+MUSE 43.7\%), representation-learning (MERIT 44.4\%), and other deep/Transformer baselines. FETA (DeepSeek R1) ranks second overall with 46.6\%, closely tracking Qwen3 and outperforming all non-FETA methods. Notably, even though FETA (LLaMA3.1) uses a smaller, non-reasoning model, it still reaches 38.9\%, matching TST (38.9\%) and clearly exceeding the direct DeepSeek R1 baseline (32.5\%) without our pipeline.

Beyond averages, \textsc{FETA} delivers strong per-dataset results. On \textit{AtrialFibrillation}, FETA (Qwen3) attains 46.7\%, far above the best baseline (33.3\%). On \textit{ERing}, FETA (LLaMA3.1) reaches 24.4\% vs. \(\sim\)13.3\% for others. On \textit{StandWalkJump}, FETA (DeepSeek R1) achieves 60.0\%, substantially outperforming TS2Vec (40.0\%) and TST (36.7\%). These results indicate that the \textsc{FETA} pipeline extracts discriminative evidence effectively, independent of model scale or specialized reasoning capabilities.

\medskip
\noindent\textbf{Why does \textsc{FETA} work?}
We attribute the gains to four complementary factors. 
\textbf{(i) Channel decomposition and selection.} By isolating and ranking channels, \textsc{FETA} preserves the most informative signals while shortening overly long sequences, preventing inflated input length that would otherwise hinder LLM reasoning efficiency. 
\textbf{(ii) Exemplar-driven in-context reasoning.} DTW-based retrieval supplies pattern-aligned neighbors, enabling each channel agent to compare the query against concrete precedents and recognize trends, shifts, and fluctuations without a learned encoder. 
\textbf{(iii) Confidence-aware decision fusion.} Requiring the LLM to output a calibrated confidence per channel and aggregating via confidence-weighted fusion enhances interpretability (per-channel rationale + certainty) and more effectively consolidates heterogeneous channel decisions than naive voting. 
\textbf{(iv) Modular multi-agent design.} By delegating distinct roles to different agents and consolidating their outputs through confidence-weighted aggregation, \textsc{FETA} effectively balances local channel-level insights with global consensus, yielding robust and interpretable classification.

\subsection{Ablation Study: Deep Dive into \textsc{FETA}’s Key Components and Mechanisms}

To assess the contribution of each component, we conduct an ablation study on three representative datasets with DeepSeek R1.

We evaluate four variants: (i) \textbf{FETA w/o Decomposer}, which removes channel decomposition/selection and feeds all channels jointly to the LLM; (ii) \textbf{FETA w/o Retriever}, which replaces DTW-based nearest-neighbor retrieval with label-wise random sampling from the training set; (iii) \textbf{FETA w/o Reasoner}, which directly adopts the top-1 exemplar’s label without LLM reasoning; and (iv) \textbf{FETA w/o Aggregator}, which replaces confidence-weighted fusion with majority voting.

As shown in Table~\ref{tab:ablation}, removing the Channel Decomposer yields the largest degradation (–14.6\% on average), with severe drops on \textit{AtrialFibrillation} and \textit{StandWalkJump} (–20.0\% each), underscoring the importance of selecting informative channels and controlling input length for efficient LLM reasoning. Eliminating DTW-based exemplar retrieval causes a notable decline (–7.6\% on average), confirming that alignment-aware neighbors are crucial for grounding in-context reasoning. Disabling the Channel Reasoner reduces accuracy by –9.7\% on average, indicating that the reasoning step is necessary to resolve label disagreements and compare temporal patterns beyond majority heuristics. Finally, removing the Decision Aggregator leads to a –9.6\% average drop, particularly pronounced on \textit{AtrialFibrillation} (–20.0\%), highlighting the benefit of confidence-aware fusion over naive voting.

Overall, these findings show that each module plays a distinct and complementary role, and their synergy under the multi-agent design enables robust, training-free classification.

\section{Conclusion}

We present \textsc{FETA}, a multi-agent framework for training-free time series classification through exemplar-based in-context reasoning. By decomposing multivariate series, retrieving DTW-aligned exemplars, reasoning at the channel level, and fusing results via confidence-weighted aggregation, \textsc{FETA} achieves efficient, interpretable, and robust classification without any training. Experiments show that \textsc{FETA} delivers competitive or superior accuracy compared to trained models, highlighting the untapped potential of LLMs as plug-and-play time series learners. This work opens a new direction for integrating reasoning LLMs with structured temporal data under zero-training constraints.

\section*{Limitations}

FETA is currently mainly designed and evaluated for multivariate time series with structured numerical signals. However, real-world applications often involve hybrid or heterogeneous data, where time series coexist with modalities such as text descriptions, images, or categorical metadata. Extending FETA’s LLM calling and modular framework to handle such multimodal or cross-domain inputs while maintaining its training-free and interpretable characteristics remains an important direction for future research.

\section*{Ethics Statement}

This work uses nine publicly available datasets from the UEA multivariate time series classification archive, which are widely adopted in prior studies and contain no personally identifiable information. No new data collection, labeling, or human annotation was conducted. Our experiments employ pretrained large language models (LLaMA 3.1, DeepSeek R1, and Qwen3) in inference-only settings without any fine-tuning or additional training. While we are not aware of dataset- or task-specific ethical concerns, we acknowledge general risks associated with pretrained LLMs, including potential propagation of social or cultural biases and the environmental cost of large-scale model deployment.

\bibliography{custom}

\appendix

\section{Dataset Details}
\label{sec:appendix_dataset}

\begin{table}[t]
\centering
\renewcommand{\arraystretch}{1.1}
\resizebox{\columnwidth}{!}{
\begin{tabular}{lccccc}
\hline
\textbf{Dataset} & \textbf{\#Classes} & \textbf{\#Channels} & \textbf{Train} & \textbf{Test} & \textbf{Length} \\
\hline
AtrialFibrillation   & 3 & 2  & 15  & 15  & 640   \\
ERing                & 6 & 4  & 30  & 270 & 65    \\
EigenWorms           & 5 & 6  & 131 & 128 & 17984 \\
EthanolConcentration & 4 & 3  & 261 & 263 & 1751  \\
FingerMovements      & 2 & 28 & 316 & 100 & 50    \\
HandMovementDirection& 4 & 10 & 160 & 74  & 400   \\
MotorImagery         & 2 & 64 & 278 & 100 & 3000  \\
SelfRegulationSCP2   & 2 & 7  & 200 & 180 & 1152  \\
StandWalkJump        & 3 & 4  & 12  & 15  & 2500  \\
\hline
\end{tabular}
}
\caption{Statistics of the nine UEA multivariate time series datasets used in our experiments.}
\label{tab:dataset_stats}
\end{table}

We evaluate FETA on nine publicly available datasets from the UEA Multivariate Time Series Classification Archive \citep{bagnall2018ueamultivariatetimeseries}.
These datasets cover diverse application domains such as biomedical signals, motion capture, and sensor-based activity recognition, with wide variation in sequence length, channel dimensionality, and number of classes, making them challenging benchmarks for training-free classification. The statistics of the datasets are shown in Table~\ref{tab:dataset_stats}.

All datasets are sourced from the official UEA repository\footnote{\url{https://www.timeseriesclassification.com}} and are freely available for research purposes under a non-commercial academic license. Each dataset includes citation and usage instructions in accordance with the original publication. No additional data collection or annotation was performed in this work.

All datasets contain numerical sensor or signal data only, with no textual, image, or personally identifiable content. The datasets are purely scientific and contain no offensive, sensitive, or privacy-related material.

\section{Implementation Details}
\label{sec:appendix_implement}

\subsection{\textsc{FETA} Framework}

All experiments are conducted using the following large language models as the backbone reasoning engines: LLaMA 3.1-8B \citep{grattafiori2024llama3herdmodels}, DeepSeek R1 \citep{deepseekai2025deepseekr1incentivizingreasoningcapability}, and Qwen 3 \citep{yang2025qwen3technicalreport}.
These models are used in inference-only mode without any additional training or fine-tuning.
Detailed model configurations are provided in Appendix~\ref{sec:appendix_parameters}, and the complete prompts used for each LLM agent are listed in Appendix~\ref{sec:appendix_prompts}.

\subsection{Tooling and Package Settings}

All experiments are implemented in Python 3.10 using standard open-source libraries.
Dynamic Time Warping (DTW) is implemented using the \texttt{fastdtw} library.
No modifications were made to third-party packages beyond configuration and parameter settings.
All experiments were executed on a Linux environment with the consistent random seed 42 to ensure reproducibility.

\section{LLMs' Inference Configurations}
\label{sec:appendix_parameters}

To ensure reproducibility and stability of outputs across all agents in our framework, we adopt consistent and deterministic decoding configurations for all backbone LLMs.

For \texttt{DeepSeek R1} model and \texttt{Qwen3} model, we use strictly deterministic inference: the \texttt{temperature} is set to 0.0 and \texttt{top\_p} to 1.0. This disables stochastic sampling and enforces greedy decoding, ensuring identical outputs across repeated runs. All responses are generated through the official APIs provided by DeepSeek and Alibaba Cloud, respectively. Reported results are obtained from a single deterministic run without sampling variability.

For \texttt{LLaMA 3.1-8B}, we use the instruct variant deployed via Hugging Face Inference Endpoints hosted on 8×A100 GPUs. Due to platform constraints, exact deterministic settings (\texttt{temperature}=0.0, \texttt{top\_p}=1.0) are not supported. As a practical approximation, we set \texttt{temperature}=0.01 and \texttt{top\_p}=0.99 for all agents, which yields nearly deterministic outputs while maintaining compatibility with the inference backend. All reported LLaMA results are likewise based on a single inference pass.

\raggedbottom

\section{Prompts of \textsc{FETA}}
\label{sec:appendix_prompts}

Our prompts are uniform across datasets and not domain-specific.

\begin{tcolorbox}[colback=gray!10, colframe=black, sharp corners=south, boxrule=0.8pt, width=\columnwidth, before skip=10pt, after skip=10pt]

\textbf{[Instruction]}\\
1. Compare the unlabeled sample ONLY with the retrieved examples shown above.

2. Focus on similarity in shape, spikes, oscillations, and recovery patterns. Ignore absolute value scale unless it clearly distinguishes classes.

3. If the majority of the retrieved examples have the same label, prioritize that label unless the sample strongly matches another.

4. Assign confidence based on neighbor consistency: All neighbors same label ~ 0.9; 2/3 neighbors same label ~ 0.7; Neighbors mixed evenly ~ 0.5

5. Return ONLY one JSON object with EXACTLY these keys and no extra text.

\medskip
\textbf{[Response format]}\\
\{
    "decision": <one of {classes}>,
    
    "confidence": <0.0 to 1.0>,
    
    "reasoning": "<one short sentence>"
    
\}

\end{tcolorbox}

\end{document}